\documentclass{article}
\usepackage{iclr2016_conference,times}
\usepackage{hyperref}
\usepackage{url}

\usepackage{graphicx} 

\usepackage{natbib}

\usepackage{amsmath, amssymb}
\usepackage{amsfonts}
\usepackage{xfrac}
\usepackage{bbm}
\usepackage{algorithm}
\usepackage{subcaption}
\usepackage{caption}

\usepackage{csquotes}
\usepackage[T1]{fontenc}

\usepackage{algorithmic}
\def\algorithmicrequire{\textbf{Input:}}
\def\algorithmicensure{\textbf{Output:}}
\def\algorithmicendfor{\textbf{end}}
\def\abovestrut#1{\rule[0in]{0in}{#1}\ignorespaces}

\def\abovespace{\abovestrut{0.20in}}

\def\algorithmiccomment#1{\textsc{#1}}

\usepackage{fixmath}
\renewcommand{\vec}[1]{\mathbold{#1}}

\title{Breaking Sticks and Ambiguities with \\ Adaptive Skip-gram}

\author{
Sergey Bartunov \\
National Research University \\
Higher School of Economics (HSE) \\
Moscow, Russia \\
\texttt{sbos@sbos.in} 
\And
Dmitry Kondrashkin \\
National Research University \\
Higher School of Economics (HSE) \\
Moscow, Russia \\
\texttt{kondra2lp@gmail.com} \\
\AND
Anton Osokin \\
INRIA~-- Sierra Project-Team, \\ 
\'{E}cole Normale Sup\'{e}rieure \\
Paris, France \\
\texttt{anton.osokin@inria.fr} 
\And
Dmitry P. Vetrov \\
National Research University \\
Higher School of Economics (HSE) \\
Moscow, Russia \\
\texttt{vetrovd@yandex.ru} \\
}

\begin{document}

\maketitle

\begin{abstract}
The recently proposed Skip-gram model is a powerful method for learning high-dimensional word representations that capture rich semantic relationships between words. However, Skip-gram as well as most prior work on learning word representations does not take into account word ambiguity and maintain only a single representation per word. Although a number of Skip-gram modifications were proposed to overcome this limitation and learn multi-prototype word representations, they either require a known number of word meanings or learn them using greedy heuristic approaches. In this paper we propose the Adaptive Skip-gram model which is a nonparametric Bayesian extension of Skip-gram capable to automatically learn the required number of representations for all words at desired semantic resolution. We derive efficient online variational learning algorithm for the model and empirically demonstrate its efficiency on word-sense induction task. 
\end{abstract}

\section{Introduction}

Continuous-valued word representations are very useful in many natural language processing applications. They could serve as input features for higher-level algorithms in text processing pipeline and help to overcome the word sparseness of natural texts. Moreover, they can explain  on their own many semantic properties and relationships between concepts represented by words.

Recently, with the success of the deep learning, new methods for learning word representations inspired by various neural architectures were introduced. Among many others the two particular models Continuous Bag of Words (CBOW) and Skip-gram (SG) proposed in \citep{SG} were used to obtain high-dimensional distributed representations that capture many semantic relationships and linguistic regularities \citep{SG,Mikolov2013b}. In addition to high quality of learned representations these models are computationally very efficient and allow to process text data in online streaming setting. 

However, word \emph{ambiguity} (which may appear as polysemy, homonymy, etc) an important property of a natural language is usually ignored in representation learning methods. For example, word ``apple'' may refer to a fruit or to the Apple inc. depending on the context. Both CBOW and SG also fail to address this issue since they assume a unique representation for each word. As a consequence either the most frequent meaning of the word dominates the others or the meanings are mixed. Clearly both situations are not desirable for practical applications.



We address the problem of unsupervised learning of multiple representations that correspond to different meanings of a word, i.e. building multi-prototype word representations. This may be considered as specific case of word sense induction (WSI) problem which consists in automatic identification of the meanings of a word. In our case different meanings are distinguished by separate representations. We define \emph{meaning} or \emph{sense} as distinguishable interpretation of the spelled word which may be caused by any kind of ambiguity.

Word-sense induction is closely related to the word-sense disambiguation (WSD) task where the goal is to choose which meaning of a word among provided in the \emph{sense inventory} was used in the context. The sense inventory may be obtained by a WSI system or provided as external information.



Many natural language processing (NLP) applications benefit from ability to deal with word ambiguity~\citep{Navigli:2010:IWS:1870658.1870670, Vickrey+al:EMNLP05}. Since word representations have been used as word features in dependency parsing~\citep{chen-manning:2014:EMNLP2014}, named-entity recognition~\citep{Turian:2010:WRS:1858681.1858721} and sentiment analysis~\citep{Maas:2011:LWV:2002472.2002491} among many other tasks, employing multi-prototype representations could increase the performance of such representation-based approaches.


In this paper we develop natural extension of the Skip-gram model which we call Adaptive Skip-gram (AdaGram). It retains all noticeable properties of SG such as fast online learning and high quality of representations while allowing to automatically learn the necessary number of prototypes per word at desired \emph{semantic resolution}.


The rest of the paper is organized as follows: we start with reviewing original Skip-gram model (section \ref{sec:SG}), then we describe our extension called Adaptive Skip-gram (section \ref{sec:DPSG}). Then, we compare our model to existing approaches in section \ref{sec:related_work}. In section \ref{sec:experiments} we evaluate our model qualitatively by considering neighborhoods of selected words in the learned latent space and by quantitative comparison against concurrent approaches. Finally, we conclude in section \ref{sec:conclusion}.

\section{Skip-gram model}\label{sec:SG}
The original Skip-gram model~\citep{SG} is formulated as a set of grouped word prediction tasks. Each task consists of prediction of a word $v$ given a word $w$ using correspondingly their \emph{output} and \emph{input} representations
\begin{equation}\label{eq:prediction}
p(v | w, \theta) = \frac{\exp(in_w^{\intercal} out_v)}{\sum_{v'=1}^V \exp(in_w^{\intercal} out_{v'})},
\end{equation}
where global parameter $\theta = \{ in_v, out_v\}_{v=1}^V$ stands for both input and output representations for all words of the dictionary indexed with $1,\dots,V$. 
Both input and output representations are real vectors of the dimensionality $D$.

These individual predictions are grouped in a way to simultaneously predict \emph{context words} $\mathbf{y}$ of some input word $x$: 
\\[-0.3pt] 
$$ p(\mathbf{y} | x, \theta) = \prod_j p(y_j | x, \theta).$$\\[-0.3pt]
Input text $\mathbf{o}$ consisting of $N$ words $o_1, o_2, \ldots, o_N$ is then interpreted as a sequence of input words $X = \{ x_i \}_{i=1}^N$ and their contexts $Y = \{ \mathbf{y}_i \}_{i=1}^N$. Here $i$-th training object ($x_i$, $\mathbf{y}_i$) consists of word $x_i=o_i$ and its context $\mathbf{y}_i = \{ o_t \}_{t \in c(i)}$ where $c(i)$ is a set of indices such that $|t - i| \leq C/2$ and $t \neq i$ for all $t \in c(i)$\footnote{For notational simplicity we will further assume that size of the context is always equal to $C$ which is true for all non-boundary words.}\!\!.

Finally, Skip-gram objective function is the likelihood of contexts given the corresponding input words:
\begin{equation}\label{eq:sg}
 p(Y | X, \theta)  = \prod_{i=1}^N p(\mathbf{y}_i | x_i, \theta) = \prod_{i=1}^N \prod_{j=1}^C p(y_{ij} | x_i, \theta).
\end{equation}
Note that although contexts of adjacent words intersect, the model assumes the corresponding prediction problems independent.

For training the Skip-gram model it is common to ignore sentence and document boundaries and to interpret the input data as a stream of words. The objective \eqref{eq:sg} is then optimized in a stochastic fashion by sampling $i$-th word and its context, estimating gradients and updating parameters~$\theta$. After the model is trained,~\cite{SG} treated the input representations of the trained model as word features and showed that they captured semantic similarity between concepts represented by the words. Further we refer to the input representations as \emph{prototypes} following \citep{Reisinger:2010:MVM:1857999.1858012}. 

Both evaluation and differentiation of~\eqref{eq:prediction} has linear complexity (in the dictionary size $V$) which is too expensive for practical applications. Because of that the soft-max prediction model \eqref{eq:prediction} is substituted by the hierarchical soft-max \citep{hierachical_softmax}:
\\[-0.3pt]\begin{equation}\label{eq:hierarchical_softmax}
    p(v | w, \theta) = \prod_{n \in path(v)} \sigma(ch(n) in_w^{\intercal} out_{n}).
\end{equation}\\[-0.3pt]
Here output representations are no longer associated with words, but rather with nodes in a binary tree where leaves are all possible words in the dictionary with unique paths from root to corresponding leaf. $ch(n)$ assigns either $1$ or $-1$ to each node in the $path(v)$ depending on whether $n$ is a left or right child of previous node in the path. Equation \eqref{eq:hierarchical_softmax} is guaranteed to sum to $1$ i.e. be a distribution w.r.t. $v$ as $\sigma(x) = 1 / (1 + \exp(-x)) = 1 - \sigma(-x)$. For computational efficiency Skip-gram uses Huffman tree to construct hierarchical soft-max. 

\section{Adaptive Skip-gram}\label{sec:DPSG}

The original Skip-gram model maintains only one prototype per word. It would be unrealistic to assume that single representation may capture the semantics of all possible word meanings. At the same time it is non-trivial to specify exactly the right number of prototypes required for handling meanings of a particular word. Hence, an adaptive approach for allocation of additional prototypes for ambiguous words is required. Further we describe our Adaptive Skip-gram (AdaGram) model which extends the original Skip-gram and may automatically learn the required number of prototypes for each word using Bayesian nonparametric approach.

First, assume that each word has $K$ meanings each associated with its own prototype. That means that we have to modify \eqref{eq:hierarchical_softmax} to account for particular choice of the meaning. For this reason we introduce latent variable $z$ that encodes the index of active meaning and extend \eqref{eq:hierarchical_softmax} to 
$p(v | z = k, w, \theta) = \prod_{n \in path(v)} \sigma(ch(n) in_{wk}^{\intercal} out_{n}).$
Note that we bring even more asymmetry between input and output representations comparing to \eqref{eq:hierarchical_softmax} since now only prototypes depend on the particular word meaning. While it is possible to make context words be also meaning-aware this would make the training process much more complicated. Our experiments show that this word prediction model is enough to capture word ambiguity. This could be viewed as prediction of context \emph{words} using \emph{meanings} of the input words.

However, setting the number of prototypes for all words equal is not a very realistic assumption. Moreover, it is desirable that the number of prototypes for a particular word would be determined by the training text corpus. We approach this problem by employing Bayesian nonparametrics into Skip-gram model, i.e. we use the constructive definition of Dirichlet process \citep{dp} for automatic determination of the required number of prototypes.  Dirichlet process (DP) has been successfully used for infinite mixture modeling and other problems where the number of structure components (e.g. clusters, latent factors, etc.) is not known a priori which is exactly our case.

We use the constructive definition of DP via the stick-breaking representation \citep{sbp} to define a prior over meanings of a word. The meaning probabilities are computed by dividing total probability mass into infinite number of diminishing pieces summing to 1. So the prior probability of $k$-th meaning of the word $w$ is \\[-12pt]
\begin{align*}
    p(z = k | w, \vec{\beta}) &= \beta_{wk} \prod_{r = 1}^{k-1}\nolimits (1 - \beta_{wr}), \quad p(\beta_{wk}|\alpha) = \mathrm{Beta}(\beta_{wk}|1, \alpha), \quad k = 1, \dots
\end{align*}
This assumes that infinite number of prototypes for each word may exist. However, as long as we consider finite amount of text data, the number of prototypes (those with non-zero prior probabilities) for word $w$ will not exceed the number of occurrences of $w$ in the text which we denote as $n_w$. The hyperparameter $\alpha$ controls the number of prototypes for a word allocated a priori. Asymptotically, the expected number of prototypes of word $w$ is proportional to $\alpha \log(n_w)$. Thus, larger values of $\alpha$ produce more prototypes which lead to more granular and specific meanings captured by learned representations and the number of prototypes scales logarithmically with number of occurrences.

Another attractive property of DPs is their ability to increase the complexity of latent variables' space with more data arriving. In our model this will result to more distinctive meanings of words discovered on larger text corpus.

Combining all parts together we may write the AdaGram model as follows:
\begin{gather*}
    p(Y, Z, \vec{\beta} | X, \alpha, \theta) = \prod_{w=1}^V \prod_{k=1}^{\infty} p(\beta_{wk} | \alpha) \prod_{i=1}^N  \Big[ p(z_i | x_i, \vec{\beta}))  \prod_{j=1}^C p(y_{ij} | z_i, x_i, \theta) \Big],
\end{gather*}
where $Z = \{z_i\}_{i=1}^N$ is a set of senses for all the words. Similarly to \cite{SG} we do not consider any regularization (and so the informative prior) for representations and seek for point estimate of $\theta$.


\subsection{Learning representations}\label{sec:learning}
One way to train the AdaGram is to maximize the marginal likelihood of the model
\begin{equation}
\label{eq:marginalLikelihood}
    \textstyle \log p(Y | X, \theta, \alpha) = \log \int \sum_{Z} p(Y, Z, \vec{\beta} | X, \alpha, \theta) d \beta 
\end{equation}
with respect to representations $\theta$.
One may see that the marginal likelihood is intractable because of the latent variables $Z$ and $\vec{\beta}$. Moreover, $\vec{\beta}$ and $\theta$ are infinite-dimensional parameters. Thus unlike the original Skip-gram and other methods for learning multiple word representations, our model could not be straightforwardly trained by stochastic gradient ascent w.r.t. $\theta$. 

To make this tractable we consider the variational lower bound on the marginal likelihood~\eqref{eq:marginalLikelihood}
\[ 
\mathcal{L} = \mathbb{E}_{q} \left[ \log p(Y, Z, \vec{\beta} | X, \alpha, \theta) - \log q(Z, \vec{\beta}) \right] 
\]
where 
$
q(Z, \vec{\beta}) = \prod_{i=1}^N q(z_i) \prod_{w=1}^V \prod_{k=1}^T q(\beta_{wk}) 
$
is the fully factorized variational approximation to the posterior $p(Z, \vec{\beta} | X, Y, \alpha, \theta)$ with possible number of representations for each word truncated to $T$~\citep{dp_varinf}. It may be shown that the maximization of the variational lower bound with respect to $q(Z,\vec{\beta})$ is equivalent to the minimization of Kullback-Leibler divergence between $q(Z, \vec{\beta})$ and the true posterior~\citep{var_inf}. 

Within this approximation the variational lower bound $\mathcal{L}(q(Z), q(\vec{\beta}), \theta)$ takes the following form:
\begingroup\makeatletter\def\f@size{7}\check@mathfonts\begin{gather*}
    \mathcal{L}(q(Z), \!q(\vec{\beta}), \!\theta) \!=\! \mathbb{E}_{q} \!\Biggl[ \sum_{w=1}^V \!\sum_{k=1}^T \!\log p(\beta_{wk} | \alpha) \!-\! \log q(\beta_{wk}) + \Bigr. 
    \Bigl. \sum_{i=1}^N \!\big( \log p(z_i | x_i, \vec{\beta}) \!-\! \log q(z_i) \!+\! \sum_{j=1}^C \log p(y_{ij} | z_i, x_i, \theta) \big) \Biggr].
\end{gather*}\endgroup

Setting derivatives of $\mathcal{L}(q(Z), q(\vec{\beta}),\theta)$ with respect to $q(Z)$ and $q(\vec{\beta})$ to zero yields standard update equations
\begin{align}  
    \log q(z_i = k) 
    &= \mathbb{E}_{q(\vec{\beta})} \Big[ \log\beta_{{x_i},k} + \sum_{r=1}^{k-1} \log (1 - \beta_{{x_i},r}) \Big] + \sum_{j=1}^C \log p(y_{ij} | k, x_i, \theta) + \text{const},
    \label{eq:updatedZ} \\
    \log q(\vec{\beta}) 
    &= \sum_{w=1}^V\nolimits \sum_{k=1}^T\nolimits \log\mathrm{Beta}(\beta_{wk}|a_{wk},b_{wk}),
    \label{eq:updatedB}
\end{align}
where (natural) parameters $a_{wk}$ and $b_{wk}$ deterministically depend on the expected number of assignments to particular sense $n_{wk} = \sum_{i : x_i = w} q(z_i = k)$ \citep{dp_varinf}: ${a_{wk} = 1 + n_{wk}}, {b_{wk} = \alpha + \sum_{r=k+1}^T n_{wr}}$.

\paragraph{Stochastic variational inference.} Although variational updates given by \eqref{eq:updatedZ} and \eqref{eq:updatedB} are tractable, they require the full pass over training data. In order to keep the efficiency of Skip-gram training procedure, we employ stochastic variational inference approach \citep{SVI} and derive online optimization algorithm for the maximization of $\mathcal{L}$. 
There are two groups of parameters in our objective: $\{ q(\beta_{vk})\}$  and $\theta$ are global because they affect all the objects; $\{ q(z_i) \}$ are local, i.e. affect only the corresponding object~$x_i$. After updating the local parameters according to \eqref{eq:updatedZ} with the global parameters fixed and defining the obtained distribution as $q^*(Z)$ we have a function of the global parameters ${\mathcal{L}^*( q(\vec{\beta}), \theta) = \mathcal{L}(q^*(Z), q(\vec{\beta}), \theta) \geq \mathcal{L}(q(Z), q(\vec{\beta}), \theta)}.$

The new lower bound $\mathcal{L}^*$ is no longer a function of local parameters which are always kept updated to their optimal values. Following \cite{SVI} we iteratively optimize $\mathcal{L}^*$ with respect to the global parameters using stochastic gradient estimated at a single object. Stochastic gradient w.r.t $\theta$ computed on the $i$-th object is computed as follows:
$$ \widehat{\nabla}_{\theta} \mathcal{L}^* = N \sum_{j=1}^C\nolimits \sum_{k=1}^T\nolimits q^*(z_i = k) \nabla_{\theta} \log p(y_{ij} | k, x_i, \theta). $$
Now we describe how to optimize $\mathcal{L}^*$ w.r.t  global posterior approximation $q(\vec{\beta}) = \prod_{w=1}^D \prod_{k=1}^T q(\beta_{wk})$. 
The stochastic gradient with respect to natural parameters $a_{wk}$ and $b_{wk}$ according to \citep{SVI} can be estimated by computing intermediate values of natural parameters $(\hat{a}_{wk}, \hat{b}_{wk})$ on the $i$-th data point as if we estimated $q(z)$ for all occurrences of $x_i = w$ equal to $q(z_i)$: 
\[
    \hat{a}_{wk} = 1 + n_w q(z_i = k), \quad \hat{b}_{wk} = \alpha + \sum_{r=k+1}^T\nolimits n_w q(z_i = r), \]
where $n_w$ is the total number of occurrences of word $w$. The stochastic gradient estimate then can be expressed in the following simple form:
$
    \widehat{\nabla}_{a_{wk}} \mathcal{L}^* = \widehat{a}_{wk} - a_{wk}, 
    \widehat{\nabla}_{b_{wk}} \mathcal{L}^* = \widehat{b}_{wk} - b_{wk}
$.
One may see that making such gradient update is equivalent to updating counts $n_{wk}$ since they are sufficient statistics of $q(\beta_{wk})$. 

We use conservative initialization strategy for $q(\vec{\beta})$ starting with only one allocated meaning for each word, i.e. $n_{w1} = n_w$ and $n_{wk} = 0, k > 1$. Representations are initialized with random values drawn from $\mathrm{Uniform}(-0.5/D, 0.5/D)$. In our experiments we updated both learning rates $\rho$ and $\lambda$ using the same linear schedule  from $0.025$ to $0$.

The resulting learning algorithm~\ref{alg:em} may be also interpreted as an instance of stochastic variational EM algorithm. It has linear computational complexity in the length of text $\mathbf{o}$ similarly to Skip-gram learning procedure. The overhead of maintaining variational distributions is negligible comparing to dealing with representations and thus training of AdaGram is $T$ times slower than Skip-gram.

\subsection{Disambiguation and prediction}\label{sec:prediction}
After model is trained on data $\mathcal{D} = \{ (x_i, \mathbf{y}_i) \}_{i=1}^N$, it can be used to infer the meanings of an input word $x$ given its context $\mathbf{y}$. The predictive probability of a meaning can be computed as
\begin{equation} \label{eq:posterior_meaning}
\textstyle p(z = k | x, \mathcal{D}, \theta, \alpha) \propto \int p(z = k | \vec{\beta}, x) q(\vec{\beta}) d \vec{\beta},
\end{equation}
where 
$
q(\vec{\beta}) 
$ 
can serve as an approximation of $p(\vec{\beta} | \mathcal{D}, \theta, \alpha)$. 
Since $q(\vec{\beta})$ has the form of independent Beta distributions whose parameters are given in sec.~3.1 the integral can be taken analytically. The number of learned prototypes for a word $w$ may be computed as $\sum_{k=1}^T \mathbbm{1} [ p(z = k | w, \mathcal{D}, \theta, \alpha) > \epsilon ]$ where $\epsilon$ is a threshold e.g. $10^{-3}$.

The probability of each meaning of $x$ given context $\mathbf{y}$ is thus given by 
\begin{equation}\label{eq:disambiguation}
\textstyle p(z = k | x, \mathbf{y}, \theta) \propto p(\mathbf{y} | x, k, \theta) \int  p(k | \vec{\beta}, x) q(\vec{\beta}) d \vec{\beta} \end{equation}

Now the posterior predictive over context words $\mathbf{y}$ given input word $x$ may be expressed as 
\begin{equation}\label{eq:predictive}
\textstyle p(\mathbf{y} | x, \mathcal{D}, \theta, \alpha) = \int \sum_{z=1}^T p(\mathbf{y} | x, z, \theta) p(z | \vec{\beta}, x) q(\vec{\beta}) d \vec{\beta}.
\end{equation}

\section{Related work}\label{sec:related_work}
Literature on learning continuous-space representations of embeddings for words is vast, therefore we concentrate on works that are most relevant to our approach.

In the works \citep{Socher2012, Reisinger:2010:MMS:1870658.1870772, Reisinger:2010:MVM:1857999.1858012} various neural network-based methods for learning multi-prototype representations are proposed. These methods include clustering contexts for all words as prepossessing or intermediate step. While this allows to learn multiple prototypes per word, clustering large number of contexts brings serious computational overhead and limit these approaches to offline setting. 

Recently various modifications of Skip-gram were proposed to learn multi-prototype representations. Proximity-Ambiguity Sensitive Skip-gram ~\citep{QiuAAAI} maintains individual representations for different parts of speech (POS) of the same word. While this may handle word ambiguity to some extent, clearly there could be many meanings even for the same part of speech of some word remaining not discovered by this approach. 

Work of ~\cite{TianColing} can be considered as a parametric form of our model with number of meanings for each word fixed. Their model also provides improvement over original Skip-gram, but it is not clear how to set the number of prototypes. Our approach not only allows to efficiently learn required number of prototypes for ambiguous words, but is able also to gradually increase the number of meanings when more data becomes available thus distinguishing between shades of same meaning. 

It is also possible to incorporate external knowledge about word meanings into Skip-gram in the form of sense inventory~\citep{chen-liu-sun:2014:EMNLP2014}. First, single-prototype representations are pre-trained with original Skip-gram. Afterwards, meanings provided by WordNet lexical database are used learn multi-prototype representations for ambiguous words. The dependency on the external high-quality linguistic resources such as WordNet makes this approach inapplicable to languages lacking such databases. In contrast, our model does not consider any form of supervision and learns the sense inventory automatically from the raw text. 

Recent work of ~\cite{NeelakantanEMNLP} proposing Multi-sense Skip-gram (MSSG) and its nonparameteric (not in the sense of Bayesian nonparametrics) version (NP MSSG) is the closest to AdaGram prior art. While MSSG defines the number of prototypes a priori similarly to \citep{TianColing}, NP MSSG features automatic discovery of multiple meanings for each word. In contrast to our approach, learning for NP MSSG is defined rather as ad-hoc greedy procedure that allocates new representation for a word if existing ones explain its context below some threshold. AdaGram instead follows more principled nonparametric Bayesian approach. 

\section{Experiments}\label{sec:experiments}

\begin{figure*}
\begin{center}
\includegraphics[width=\textwidth]{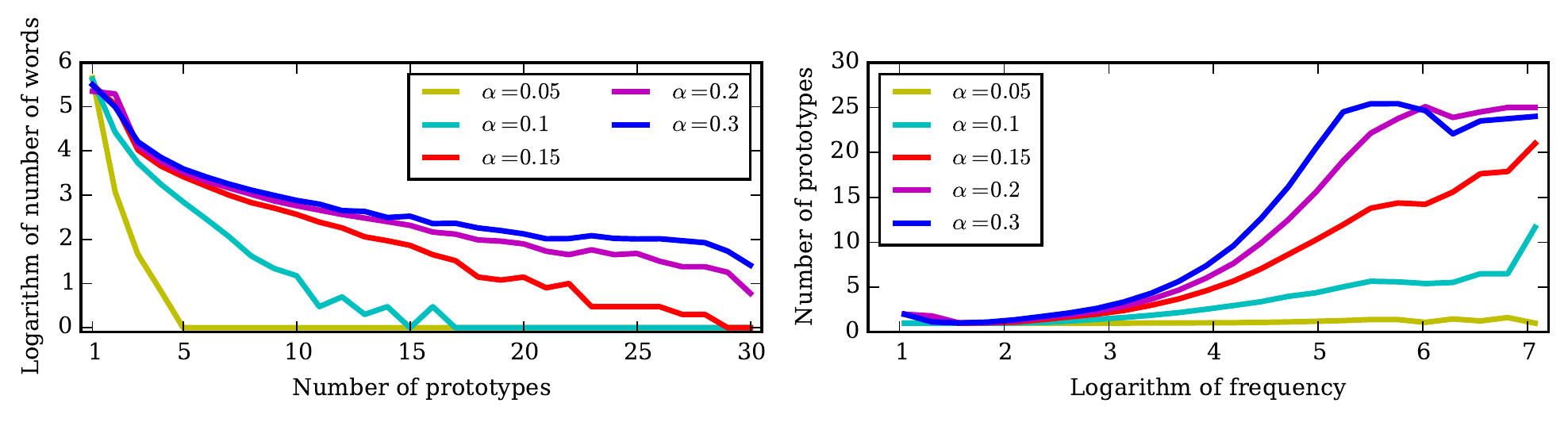}
\caption{Left: Distribution of number of word meanings learned by
AdaGram model for different values of parameter $\alpha$.
For the number of meanings $k$ we plot the $\log_{10}(n_k+1)$,
where $n_k$ is the number of words with $k$ meanings.
Right: All words in dictionary were divided into 30 bins according to the logarithm of their frequency.
Here we plot the number of learned prototypes averaged over each such bin. }
\vspace{-0.5cm}
\label{fig:SR}
\end{center}
\end{figure*}

In this section we empirically evaluate our model in a number of different tests. 
First, we demonstrate learned multi-prototype representations on several example words. 
We investigate how different values of $\alpha$ affect the number of learned prototypes what we call a semantic resolution of a model. Then we evaluate our approach on the word sense induction task (WSI). We also provide more experiments in the supplementary material. 

In order to evaluate our method we trained several models with different values of $\alpha$ on April~2010 snapshot of English Wikipedia~\citep{Westbury}. It contains nearly~2~million articles and~990 million tokens. We did not consider words which have less than~20 occurrences. The context width was set to $C=10$ and the truncation level of Stick-breaking approximation (the maximum number of meanings) to~$T=30$.
The dimensionality $D$ of representations learned by our model was set to $300$ to match the dimensionality of the models we compare with.


\subsection{Nearest neighbours of learned prototypes}

In Table~\ref{table:NN} (see appendix) we present the meanings which were discovered by our model with parameter $\alpha=0.1$ for words used in \citep{NeelakantanEMNLP}  and for a few other sample words. 
To distinguish the meanings we obtain their nearest neighbors by computing the cosine similarity
between each meaning prototype and the prototypes of meanings of all other words.
One may see that AdaGram model learns a reasonable number of prototypes which are meaningful and interpretable.
The predictive probability of each meaning reflects how frequently it was used in the training corpus. 

For most of the words $\alpha=0.1$ results in most interpretable model. It seems that for values less than $0.1$ for most words only one prototype is learned and for values greater than $0.1$ the model becomes less interpretable as learned meanings are too specific sometimes duplicating.

\subsection{Semantic resolution}

\begin{table}[t]
\caption{Nearest neighbours of different prototypes of words ``light'' and ``core'' learned by AdaGram under different values of $\alpha$ and corresponding predictive probabilities.}\vspace{-0.4cm}
\label{tbl:resolution}
\begin{center}
\begin{small}
\begin{tabular}{cll}
\multicolumn{1}{c}{\rule{0pt}{0.9\normalbaselineskip}\bf ALPHA} & \multicolumn{1}{c}{\bf ``LIGHT''} & \multicolumn{1}{c}{\bf ``CORE''} \\
 & \begin{tabular}{cc}
 $p(z)$ & nearest neighbours
\end{tabular} & \begin{tabular}{cc}
 $p(z)$ & nearest neighbours
\end{tabular}
\\ \hline \\ 
Skip-gram & \begin{tabular}{cl}
 1.00 & far-red, emitting
\end{tabular} & \begin{tabular}{cc}
 1.00 & cores, component, i7
\end{tabular} \\
0.05 & \begin{tabular}{cl}
 1.00 & far-red, illumination
\end{tabular} & \begin{tabular}{cl}
 0.40 & corium, cores, sub-critical \\
 0.60 & basic, i7, standards-based \\
\end{tabular} \\
0.075 & \begin{tabular}{cl}
 0.28 & armoured, amx-13, kilcrease \\
 0.72 & bright, sunlight, luminous
\end{tabular} & \begin{tabular}{cl}
0.30 & competencies, curriculum \\
0.34 & cpu, cores, i7, powerxcell \\
0.36 & nucleus, backbone
\end{tabular} \\ 
0.1 & \begin{tabular}{cl}
0.09 & tv\"{a}rbanan, hudson-bergen \\
0.17 & dark, bright, green \\
0.09 & 4th, dragoons, 2nd \\
0.26 & radiation, ultraviolet \\
0.28 & darkness, shining, shadows \\
0.11 & self-propelled, armored 
\end{tabular} & \begin{tabular}{cl}
0.21 & reactor, hydrogen-rich \\
0.13 & intel, processors \\
0.27 & curricular, competencies \\
0.15 & downtown, cores, center \\
 0.24 & nucleus, rag-tag, roster 
\end{tabular}
\end{tabular} 
\end{small}
\vspace{-0.4cm}
\end{center}
\end{table}

As mentioned in section~\ref{sec:DPSG} hyperparameter~$\alpha$ of the AdaGram model indirectly controls the number of induced word meanings. Figure~\ref{fig:SR}, Left shows the distribution of
number of induced word meanings under different values of $\alpha$. One may see that while for most words relatively small number of meanings is learned, larger values of $\alpha$ lead to more meanings in general. This effect may be explained by the property of Dirichlet process to allocate number of prototypes that logarithmically depends on number of word occurrences. Since word occurrences are known to be distributed by Zipf's law, the majority of words is rather infrequent and thus our model discovers few meanings for them. Figure~\ref{fig:SR}, Right quantitatively demonstrates this phenomenon. 

In the Table \ref{tbl:resolution} we demonstrate how larger values of $\alpha$ lead to more meanings on the example of the word ``light''. The original Skip-gram discovered only the meaning related to a physical phenomenon, AdaGram with $\alpha=0.075$ found the second, military meaning, with further increase of $\alpha$ value those meanings start splitting to submeanings, e.g. light tanks and light troops.
Similar results are provided for the word ``core''. 

\subsection{Word prediction}\label{sec:word_prediction}
\begin{table}[t]
\caption{Test log-likelihood under different $\alpha$ on sample from Wikipedia (see sec.~\ref{sec:word_prediction}) and Adjusted Rand Index (ARI) on the training part of
WWSI dataset (see sec.~\ref{sec:wsi}).}
\label{tbl:likelihood}
\vspace{-0.3cm}
\begin{center}
\begin{small}
\begin{tabular}{lcc}
\multicolumn{1}{c}{\bf \rule{0pt}{0.9\normalbaselineskip}MODEL}  &\multicolumn{1}{c}{\bf LOG-LIKELIHOOD}&\multicolumn{1}{c}{\bf ARI}\\
\hline \\
Skip-Gram.300D & -7.403 & -\\
Skip-Gram.600D & -7.387 & -\\
AdaGram.300D $\alpha=0.05$ & -7.399 & 0.007\\
AdaGram.300D $\alpha=0.1$ & -7.385  & 0.226\\
AdaGram.300D $\alpha=0.15$ & -7.382 & \bf{0.268}\\
AdaGram.300D $\alpha=0.2$ & -7.378  & 0.254\\
AdaGram.300D $\alpha=0.25$ & \bf -7.375 & 0.250 \\
AdaGram.300D $\alpha=0.5$ & -7.387      & 0.230 \\
\end{tabular}
\end{small}
\vspace{-0.5cm}
\end{center}
\end{table}

Since both Skip-gram and AdaGram are defined as models for predicting context of a word, it is essential to evaluate how well they explain test data by predictive likelihood. We use last 200 megabytes of December 2014 snapshot of English Wikipedia as test data for this experiment.

Similarly to the train procedure we consider this text as pairs of input words and contexts of size $C=10$, that is, $\mathcal{D}_{test} = \{ (x_i, \mathbf{y}_i) \}_{i=1}^N$ and compare AdaGram with original Skip-gram by average log-likelihood (see sec.~\ref{sec:prediction}). We were unable to include MSSG and NP-MSSG into the comparison as these models do not estimate conditional word likelihood. The results are given in Table \ref{tbl:likelihood}. Clearly, AdaGram models text data better than Skip-gram under wide range of values of $\alpha$.

Since AdaGram has more parameters than Skip-Gram with the same dimensionality of representations, it is natural to compare its efficiency with Skip-Gram that has the comparable number of parameters. The model with $\alpha=0.15$ which we study extensively further has approximately $2$ learned prototypes per word in average, so we doubled the dimensionality of Skip-Gram and included it into comparison as well\footnote{Also note that $600$-dimensional Skip-Gram has twice more parameters in hierarchical softmax than $300$-dimensional AdaGram}. One may see that AdaGram with $\alpha$ equal to $0.15$ outperforms $600$-dimensional Skip-Gram and so does the model with $\alpha=0.1$.

\begin{figure*}[t]
\begin{center}
\includegraphics[width=\textwidth, clip=true, trim=0cm 0.2cm 0cm 0.4cm]{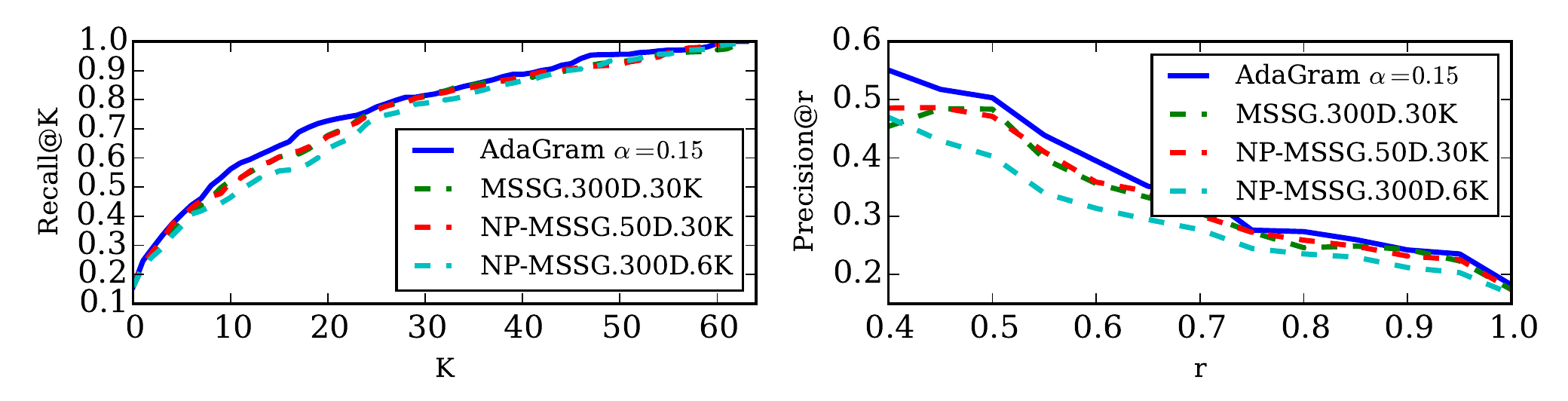}
\caption{Recall (left) and precision (right) for task-11 of Semeval-2013 competition. 
$K$ is the position of the snippet in the search result page,
$r$ is the recall value, 
see text for details.}
\label{fig:semeval-2013-11}
\end{center}
\vspace{-0.5cm}
\end{figure*}

\subsection{Word-sense induction}\label{sec:wsi}

\begin{table*}[t]
\caption{Adjusted rand index (ARI) for word sense
induction task for different datasets. Here we use the test
subset of WWSI dataset. See sec.~\ref{sec:wsi} for details. }
\vspace{-0.3cm}
\label{table:ari-all}
\begin{center}
\begin{footnotesize}
\begin{tabular}{lc@{}c@{}c@{}c}
\multicolumn{1}{c}{\bf \rule{0pt}{0.9\normalbaselineskip} MODEL}  &\multicolumn{1}{c}{\bf SE-2007} & \multicolumn{1}{c}{\bf SE-2010}  & \multicolumn{1}{c}{\bf SE-2013} & \multicolumn{1}{c}{\bf WWSI}
\\ \hline \\
MSSG.300D.30K & 0.048& 0.085& 0.033 & 0.194\\
NP-MSSG.50D.30K & 0.031& 0.058& 0.023 & 0.163\\
NP-MSSG.300D.6K & 0.033& 0.044& 0.033 & 0.110\\
MPSG.300D & 0.044 & 0.077 & 0.014 & 0.160 \\
AdaGram.300D.$\alpha$\,$=$\,$0.15$
& \bf{0.069}& \bf{0.097}& \bf{0.061}& \bf{0.286}\\
\end{tabular}
\end{footnotesize}
\vspace{-0.5cm}
\end{center}
\end{table*}

The nonparametric learning of a multi-prototype representation model is closely related to the word-sense induction (WSI) task which aims at automatic discovery of different meanings for the words. Indeed, learned prototypes identify different word meanings and it is natural to assess how well they are aligned with human judgements. 

We compare our AdaGram model with Nonparametric Multi-sense Skip-gram (NP-MSSG) proposed by \citet{NeelakantanEMNLP} which is currently the only existing approach to learning multi-prototype word representations with Skip-gram. We also include in comparison the parametric form of NP-MSSG which has the number of meanings fixed to 3 for all words during the training. All models were trained on the same dataset which is the Wikipedia snapshot by~\citet{Westbury}. For the comparison with MSSG and NP-MSSG we used source code and models released by the authors. \citet{NeelakantanEMNLP} limited the number of words for which multi-prototype representations were learned (30000 and 6000 most frequent words) for these models. We use the following notation: 300D or 50D is the dimensionality of word representations, 6K or 30K is the number of multi-prototype words  (6000 and 30000 respectively)
in case of MSSG and NP-MSSG models. Another baseline is Multi-prototype Skip-Gram (MPSG) proposed by \cite{TianColing} which can be seen as a special case of AdaGram with number of senses fixed. We have trained this model similarly to \citep{TianColing} setting number of senses for each word equal to 3.

The evaluation is performed as follows. Dataset consisting of \emph{target word} and \emph{context} pairs is supplied to a model which uses the context to disambiguate target word into a meaning from its learned sense inventory. Then for each target word the model's labeling of contexts and ground truth one are compared as two different clusterings of the same set using appropriate metrics. The results are then averaged over all target words.
\paragraph{Data}
We consider several WSI datasets in our experiments. The SemEval-2007 dataset was introduced for SemEval-2007 Task 2 competition, it contains 27232 contexts collected from Wall Street Journal (WSJ) corpus. The SemEval-2010 was similarly collected for the SemEval-2010 Task 14 competition and contains 8915 contexts in total, part obtained from web pages returned by a search engine and the other part from news articles. We also consider SemEval-2013 Task 13 dataset consisting from 4664 contexts (we considered only single-term words in this dataset).

In order to make the evaluation more comprehensive, we introduce the new Wikipedia Word-sense Induction (WWSI) dataset consisting of 188 target words and 36354 contexts. For the best of our knowledge it is currently the largest WSI dataset available. While SemEval datasets are prepared with hand effort of experts which mapped contexts into gold standard sense inventory, we collected WWSI using fully automatic approach from December 2014 snapshot of Wikipedia. The dataset is splitted evenly into train and test parts. More details on the dataset construction procedure are provided in the supplementary material, sec. 3.

For all SemEval datasets we merged together train and test contexts and used them for model comparison. Each model was supplied with contexts of the size that maximizes its ARI performance.

\paragraph{Metrics}
Authors of SemEval dataset~\citet{manandhar2010semeval} suggested two metrics for model comparison: V-Measure~(VM) and F-Score~(FS). They pointed to the weakness of both VM and FS. VM favours large number of clusters and attains large values on unreasonable clusterings which assign each instance to its own cluster while FS is biased towards clusterings consisting of small number of clusters e.g. assigning each instance to the same single cluster. Thus we consider another metric - adjusted Rand index (ARI) \citep{hubert1985comparing} which does not suffer from such drawbacks. Both examples of undesirable clusterings described above will get ARI of nearly zero which corresponds to human intuition. Thus we consider ARI as more reliable metric for WSI evaluation. We still report VM and FS values in the suppl. material (sec. 4) in order to make our results comparable to others obtained on the datasets.
\paragraph{Evaluation} 
Since AdaGram is essentially influenced by hyperparameter $\alpha$, we first investigate how different choices of $\alpha$ affect WSI performance on the train part of our WWSI dataset in terms of ARI, see table~\ref{tbl:likelihood}. The model with $\alpha=0.15$ attains maximum ARI and thus we use this model for all further experiments.

We compare AdaGram against MSSG and NP-MSSG on all datasets described above, see table~\ref{table:ari-all} for results. AdaGram consistently outperforms the concurrent approaches on all datasets and achieves significant improvement on the test part of WWSI dataset. One may see that nonparametric version of MSSG delivers consistently worse performance than MSSG with number of prototypes fixed to 3. This suggests that the ability to discover different word meanings is rather limited for NP-MSSG. The fact that AdaGram substantially outperforms NP-MSSG indicates that more principled Bayesian nonparametric approach is more suitable for the task of word-sense induction.

One may note that results on SemEval datasets are smaller than results on WWSI dataset consistently for all models. We explain this by the difference between the train corpus and test data used for preparing SemEval datasets such as news articles as well as by the difference between sense inventories, i.e. SemEval data uses WordNet and OntoNotes as sources of word meanings.



\subsection{Web search results diversification}
In this experiment we follow the methodology of Semeval-2013 Task 11 competition.
Systems are given an ambiguous query and web search result
snippets which have to be clustered.
The main goal of this task is to measure the ability of
systems to diversify web search results.
The authors of this task~\citep{di2013clustering} proposed to use two following metrics for evalutation.
Subtopic Recall@K measures how many different word meanings (from the gold standard sense inventory) are covered by top $K$ diversified search results. 
Subtopic Precision@r determines the ratio of different meanings provided in the first $K_r$ results where $K_r$ is minumum number of top $K$ results achieving recall $r$.
We consider only single-token words in this comparison.
The results are shown in the Figure~\ref{fig:semeval-2013-11}. One may see that curves of AdaGram are monotonically higher than curves of all concurrent models suggesting that AdaGram is more suitable on such real-world application.

\section{Conclusion}\label{sec:conclusion}

In the paper we proposed AdaGram which is the Bayesian nonparametric extension of the well-known Skip-gram model. AdaGram uses different prototypes to represent a word depending on the context and thus may handle various forms of word ambiguity. Our experiments suggest that representations learned by our model correspond to different word meanings. Using resolution parameter $\alpha$ we may control how many prototypes are extracted from the same text corpus. Too large values of $\alpha$ lead to different prototypes that correspond to the same meaning which decreases model performance. The values $\alpha=0.1 - 0.2$ are generally good for practical purposes. For those values the truncation level $T=30$ is enough and does not affect the number of discovered prototypes. AdaGram also features online variational learning algorithm which is very scalable and makes it possible to train our model just several times slower than extremely efficient Skip-gram model. Since the problem of learning multi-prototype word representation is closely related to word-sense induction, we evaluated AdaGram on several WSI datasets and contributed a new large one obtained automatically from Wikipedia disambiguation pages. The source code of our implementation, WWSI dataset and all trained models are available at \url{http://github.com/sbos/AdaGram.jl}.  

\bibliographystyle{iclr2016_conference}
\bibliography{references}

\newpage

\section{Appendix}

\begin{algorithm}
   \caption{Training AdaGram model \label{alg:learning}}
   \label{alg:em}
\begin{algorithmic}
   \REQUIRE training data $\{ (x_i, \mathbf{y}_i) \}_{i=1}^N$, hyperparameter $\alpha$ 
   \ENSURE parameters $\theta$, distributions $q(\vec{\beta})$, $q(\vec{z})$
   \STATE Initialize parameters $\theta$, distributions $q(\vec{\beta})$, $q(\vec{z})$
   \FOR{$i=1$ {\bfseries to} $N$}
   \STATE Select word $w = x_i$ and its context $\vec{y}_i$
    \item[]\COMMENT{Local step:}
    \FOR{$k=1$ {\bfseries to} $T$}
    \STATE $\gamma_{ik} = \mathbb{E}_{q(\beta_w)} [ \log p(z_i = k | \vec{\beta}, x_i) ]$
        \FOR{$j=1$ {\bfseries to} $C$}
            \STATE $\gamma_{ik} \gets \gamma_{ik} + \log p(y_{ij} | x_i, k, \theta)$
        \ENDFOR
   \ENDFOR
   	\STATE $\gamma_{ik} \gets \exp(\gamma_{ik}) / \sum_\ell \exp(\gamma_{i\ell}) $  
   	\item[] \COMMENT{Global step:}
    \STATE $\rho_t \gets 0.025 (1 - i/N)$, $\lambda_t \gets 0.025 (1 - i/N)$
    \FOR{$k=1$ {\bfseries to} $T$}
        \STATE Update $n_{wk} \gets (1 - \lambda_t) n_{wk} + \lambda_t n_{w} \gamma_{ik}$
    \ENDFOR
   	\STATE Update $\theta \gets \theta + \rho_t \nabla_{\theta} \sum_k \sum_j \gamma_{ik} \log p(y_{ij} | x_i, k, \theta)$
   \ENDFOR
\end{algorithmic}
\end{algorithm}

\begin{table}[t]
\caption{Nearest neighbors of meaning prototypes learned by
the AdaGram model with $\alpha=0.1$.
In the second column we provide the predictive
probability of each meaning.}
\label{table:NN}
\begin{center}
\begin{small}
\begin{tabular}{lcl}
\bf WORD & $p(z)$ & \bf NEAREST NEIGHBOURS \\
\hline \\
python
& 0.33 & monty, spamalot, cantsin \\
& 0.42 & perl, php, java, c++ \\
& 0.25 & molurus, pythons \\
apple
& 0.34 & almond, cherry, plum \\
& 0.66 & macintosh, iifx, iigs \\
date
& 0.10 & unknown, birth, birthdate \\
& 0.28 & dating, dates, dated \\
& 0.31 & to-date, stateside \\
& 0.31 & deadline, expiry, dates \\
bow
& 0.46 & stern, amidships, bowsprit \\
& 0.38 & spear, bows, wow, sword \\
& 0.16 & teign, coxs, evenlode \\
mass
& 0.22 & vespers, masses, liturgy \\
& 0.42 & energy, density, particle \\
& 0.36 & wholesale, widespread \\
run
& 0.02 & earned, saves, era \\
& 0.35 & managed, serviced \\
& 0.26 & 2-run, ninth-inning \\
& 0.37 & drive, go, running, walk \\
net
& 0.34 & pre-tax, pretax, billion \\
& 0.28 & negligible, total, gain \\
& 0.16 & fox, est/edt, sports \\
& 0.23 & puck, ball, lobbed \\
fox
& 0.38 & cbs, abc, nbc, espn \\
& 0.14 & raccoon, wolf, deer, foxes \\
& 0.33 & abc, tv, wonderfalls \\
& 0.14 & gardner, wright, taylor \\
rock
& 0.23 & band, post-hardcore \\
& 0.10 & little, big, arkansas \\
& 0.29 & pop, funk, r\&b, metal, jazz \\
& 0.14 & limestone, bedrock \\
& 0.23 & 'n', roll, `n', 'n 
\end{tabular}
\end{small}
\end{center}
\end{table}

\subsection{Full evaluation on WSI task}

We report V-measure and F-score values in tables~\ref{table:vmeasure-all} and \ref{table:fscore-all}. The results are rather contradicting due to the reasons we described in the paper: while V-measure prefers larger number of meanings F-score encourages small number of meanings. We report these numbers in order to make the values comparable with other results.

\subsection{Full evaluation of SemEval-2013 Task-13}

This task evaluates Word Sense Induction systems by performing
fuzzy clustering comparison, i.e. in the gold standard each context could be assigned to several meanings with some score indicating confidence of the assignment. Two metrics were used for comparing such fuzzy clusterings:
Fuzzy Normalized Mutual Information and Fuzzy-B-Cubed which are
introduced in~\citep{jurgens2013semeval}. Fuzzy-NMI measures the alignment of
two clustering and it is independent of the cluster sizes.
It is suitable to measure how well the model captures rare senses.
On the contrary Fuzzy-B-Cubed is sensitive to the cluster sizes. So it reflects the performance of the system on a dataset where the clusters have almost the same frequency. Results of MSSG, NP-MSSG and AdaGram models are shown in Table~\ref{table:semeval-2013-13}.

However these measures have similar drawbacks as V-Measure and F-score described above.
Trivial solution like assigning one sense per each context obtains high value of Fuzzy-NMI while treating each  word as single-sense one performs well in terms of Fuzzy-B-Cubed. All WSI systems participated in this task failed to completely surpass these baselines according to the Table~3 in~\citep{jurgens2013semeval}. Hence we consider ARI comparison as more reliable. Note that since we excluded multi-token words from the evaluation the numbers we report are not comparable with other results made on the dataset.

The ARI comparison we report in the paper was done by transforming fuzzy clusterings into hard ones, i.e. each context was assigned to most probable meaning. 


\subsection{WWSI Dataset construction details}

Similarly to \citep{navigli-vannella:2013:SemEval-2013} we considered Wikipedia's disambiguation pages as a list of ambiguous words. From that list we have selected target single-term words which had occurred in the text at least 5000 times to ensure there is enough training contexts in Wikipedia to capture different meanings of a word (note, however, that all models were trained on earlier snapshot of Wikipedia). We also did not consider pages belonging to some categories such as ``Letter-number\_combination\_disambiguation\_pages'' as they did not contain meaningful words. Then we prepared the sense inventory for each word in the list using Wikipedia pages with names matching to the pattern ``WORD\_(*)'' which is used as convenient naming of specific word meanings. Again, we applied some automatic filtering to remove names of people and geographical places in order to obtain more coarse-grained meanings. Finally for each page selected on the previous step we find all occurrences of the target word on it and use its 5-word neighbourhood (5 words on the left and 5 words on the right) as a context. Such size of the context was chosen to minimize the intersection between adjacent contexts but still provide enough words for disambiguation. 10-word context results into average intersection of $1.115$ words.

The list of the categories pages belonging to which were excluded during target word selection is following:
\begin{itemize}
\item Place\_name\_disambiguation\_pages
\item Disambiguation\_pages\_with\_surname-holder\_lists
\item Human\_name\_disambiguation\_pages
\item Lists\_of\_ambiguous\_numbers
\item Disambiguation\_pages\_with\_given-name-holder\_lists
\item Letter-number\_combination\_disambiguation\_pages
\item Two-letter\_disambiguation\_pages
\item Transport\_route\_disambiguation\_pages
\item Temple\_name\_disambiguation\_pages
\item and also those from the categories which name contains one of the substrings: ``cleanup'', ``people'', ``surnames''
\end{itemize}

During the sense inventory collection we do not consider pages which name contains one of the following substrings: ``tv\_'', ``series'', ``movie'', ``film'', ``song'', ``album'', ``band'', ``singer'', ``musical", ``comics"; and also those from the categories with names containing geography terms ``countries'', ``people'', ``province'', ``provinces''.

\section{Experiments on contextual word similarity}
In this section we compare AdaGram to other multi-prototype models on the contextual word similarity task using the SCWS dataset proposed in \citep{Socher2012}. The dataset consists of 2003 pairs of words each assigned with 10 human judgements on their semantic similarity. The common evaluation methodology is to average these 10 values for each pair and measure Spearman's rank correlation of the result and the similarities obtained using word representations learned by a model, i.e.\ by a cosine similarity of corresponding vectors. 

There are two measures of word similarity based on context: expected similarity of prototypes with respect to posterior distributions given contexts
\begin{small}
\begin{gather*}
    AvgSimC(w_1, w_2) = \frac{1}{K_1} \frac{1}{K_2} \cdot \\ \sum_{k_1} \sum_{k_2} p(k_1 | w_1, C_1) p(k_2 | w_2, C_2) \cos(vec(w_1, k_1), vec(w_2, k_2)),
\end{gather*}
\end{small}
and similarity of the most probable prototypes given contexts
\[
    MaxSimC(w_1, w_2) = \cos(vec(w_1, k_1), vec(w_2, k_2)),
\]
where $k1 = \arg\max_{k} p(k | w_1, C_1)$ and $k2 = \arg\max_{k} p(k | w_2, C_2)$, correspondingly. Here we define $K_1$ and $K_2$ as the number learned prototypes for each of the words and $C_1$, $C_2$ as their corresponding contexts. In AdaGram $vec(w, k) = In_{wk}$ and the posterior distribution over word senses is computed according to sec.~3.2. For word disambiguation AdaGram uses 4 nearest words in a context. Results for NP-MSSG and MSSG are taken from~\citep{NeelakantanEMNLP} and results for MPSG~-- from~\citep{TianColing}.

We also consider the original Skip-gram model as a baseline. We train two models: the first one with prototypes of dimensionality $300$ based on hierarchical soft-max and the second one of dimensionality~$900$ trained using negative sampling~\citep{SG} (number of negative samples is set to 5, three iterations over training data are made). The training data and other parameters are identical to the training of AdaGram used in main experiments. Note that for Skip-gram measures $AvgSimC$ and $MaxSimC$ coincide because the model learns only one representation per word.

The results on the experiment are provided in table~\ref{table:SCWS}. NP-MSSG model of \cite{NeelakantanEMNLP} outperforms other models in terms of $AvgSimC$, however, one may see that the improvement over 900-dimensional Skip-Gram baseline is only marginal, moreover, the latter is the second best model despite ignoring the contextual information and hence being unable to distinguish between different word meanings. This may suggest that SCWS is of limited use for evaluating multi-prototype word representation models as the ability of differentiating between word senses is not necessary to achieve a good score. One may  consider another example of an undesirable model which will not be penalized by the target metric in the world similarity task. That is, if a model learned too many prototypes for a word, e.g. with very close vector representations it is hardly usable in practice, but as long as averaged similarities between prototypes correlate with human judgements such non-interpretability will not be accounted during evaluation. We thus consider word-sense induction as a more natural task for evaluation since it explicitly accounts for proper and interpretable mapping from contexts into discovered word meanings. 

\begin{table*}[t]
\caption{V-Measure for word sense
induction task for different datasets. Here we use the test
subset of WWSI dataset.}
\label{table:vmeasure-all}
\begin{center}
\begin{tabular}{lcccc}
\multicolumn{1}{c}{\bf MODEL}  &\multicolumn{1}{c}{\bf SEMEVAL-2007} & \multicolumn{1}{c}{\bf SEMEVAL-2010}  & \multicolumn{1}{c}{\bf SEMEVAL-2013} & \multicolumn{1}{c}{\bf WWSI}
\\ \hline \\
MSSG.300D.30K                       & 0.067 & 0.144 & 0.033 & 0.215 \\
NP-MSSG.50D.30K                     & 0.057 & 0.119 & 0.023 & 0.188 \\
NP-MSSG.300D.6K                     & 0.073 & 0.089 & 0.033 & 0.128 \\
AdaGram.300D $\alpha$\,$=$\,$0.15$  & \bf{0.114 }& \bf{0.200}& \bf{0.192} & \bf{0.326} \\
\end{tabular}
\end{center}
\end{table*}

\begin{table*}[t]
\caption{F-Score for word sense
induction task for different datasets. Here we use the test
subset of WWSI dataset.}
\label{table:fscore-all}
\begin{center}
\begin{tabular}{lcccc}
\multicolumn{1}{c}{\bf MODEL}  &\multicolumn{1}{c}{\bf SEMEVAL-2007} & \multicolumn{1}{c}{\bf SEMEVAL-2010}  & \multicolumn{1}{c}{\bf SEMEVAL-2013} & \multicolumn{1}{c}{\bf WWSI}
\\ \hline \\
MSSG.300D.30K                       & 0.528 & 0.492 & \bf{0.437}& 0.632\\
NP-MSSG.50D.30K                     & 0.496 & 0.488 & 0.392 & 0.621 \\
NP-MSSG.300D.6K                     & \bf{0.557}& \bf{0.531}& 0.419 & \bf{0.660}\\
AdaGram.300D $\alpha$\,$=$\,$0.15$  & 0.448 & 0.439 & 0.342 & 0.588 \\
\end{tabular}
\end{center}
\end{table*}

\begin{table*}[t]
\caption{Fuzzy Normalized Mutual Information and Fuzzy B-Cubed
metric values for task-13 of Semeval-2013 competition. See text for details.}
\label{table:semeval-2013-13}
\begin{center}
\begin{tabular}{lcc}
\multicolumn{1}{c}{\bf MODEL}  &\multicolumn{1}{c}{\bf FUZZY-NMI} &\multicolumn{1}{c}{\bf FUZZY-B-CUBED}
\\ \hline \\
MSSG.300D.30K   & 0.070 & 0.287\\
NP-MSSG.50D.30K & 0.064 & 0.273\\
NP-MSSG.300D.6K & 0.063 & \bf{0.290}\\
AdaGram.300D $\alpha$\,$=$\,$0.15$ & \bf{0.089} & 0.132 \\
\end{tabular}
\end{center}
\end{table*}

\begin{table*}[t]
\caption{Spearman's rank correlation results for contextual similarity task on SCWS dataset. Numbers are multiplied with 100.}
\label{table:SCWS}
\begin{center}
\begin{tabular}{lcc}
\multicolumn{1}{c}{\bf MODEL}  &\multicolumn{1}{c}{\bf $AvgSimC$} &\multicolumn{1}{c}{\bf $MaxSimC$}
\\ \hline \abovespace
MSSG.300D.30K   & \bf 69.3 & 57.26 \\
NP-MSSG.50D.30K & 66.1 & 50.27 \\
NP-MSSG.300D.6K & 69.1 & 59.8 \\
MPSG.300D & 65.4 & 63.6 \\
Skip-Gram.300D & 65.2 & 65.2 \\
Skip-Gram.900D & 68.4 & \bf 68.4 \\
AdaGram.300D $\alpha$\,$=$\,$0.15$ & 61.2 & 53.8 
\end{tabular}
\end{center}
\end{table*}

\end{document}